\pdfoutput=1
\documentclass[11pt]{article}

\usepackage{acl}

\usepackage{times}
\usepackage{latexsym}
\usepackage{times}
\usepackage{latexsym}
\usepackage{multirow}
\usepackage{adjustbox}
\usepackage{amssymb}
\usepackage{enumitem}
\usepackage{graphicx}
\usepackage{amsmath}
\usepackage{amsthm}
\usepackage{booktabs}
\usepackage{stfloats}
\usepackage[T1]{fontenc}
\usepackage[utf8]{inputenc}

\usepackage{microtype}

\usepackage{natbib}
\usepackage{amsmath}
\usepackage{amsfonts}
\usepackage{pifont}
\usepackage{adjustbox}

\title{Adopting the Multi-answer Questioning Task with an Auxiliary Metric for Extreme Multi-label Text Classification Utilizing the Label Hierarchy}

\author{Li Wang$^{1}$, Ying Wah Teh$^{1}$\thanks{corresponding author}  , Mohammed Ali Al-Garadi$^{2}$\\
	$^{1}$Faculty of Computer Science and Information Technology,  University of Malaya\\
	$^{2}$Vanderbilt University Medical Center \\
	$^{1}${\tt wva180023@um.edu.my, tehyw@um.edu.my} \\
	$^{2}${\tt mohammed.a.al-garadi@vumc.org} \\
}
\begin{document}
	\maketitle
	\begin{abstract}
	Extreme multi-label text classification utilizes the label hierarchy to partition extreme labels into multiple label groups, turning the task into simple multi-group multi-label classification tasks. Current research encodes labels as a vector with fixed length which needs establish multiple classifiers for different label groups. The problem is how to build only one classifier without sacrificing the label relationship in the hierarchy. This paper adopts the multi-answer questioning task for extreme multi-label classification. This paper also proposes an auxiliary classification evaluation metric. This study adopts the proposed method and the evaluation metric to the legal domain. The utilization of legal Berts and the study on task distribution are discussed. The experiment results show that the proposed hierarchy and multi-answer questioning task can do extreme multi-label classification for EURLEX dataset. And in minor/fine-tuning the multi-label classification task, the domain adapted BERT models could not show apparent advantages in this experiment. The method is also theoretically applicable to zero-shot learning. 

	\end{abstract}
	
	\section{Introduction}
	
	Extreme Multi-Label Text Classification (XMTC) is a persistent problem in big data mining. It selects labels for a text from tens of thousands label collections. The difficulty of this task lies in the extremely severe data sparsity, few or zero shot learning, the difficulty in learning the dependency patterns between labels, and the high computational cost of training and testing \footnote{\url{https://analyticsindiamag.com/what-is-extreme-multilabel-text-classification/}}. The approaches dealt with by XMTC are One-Vs-All (OVA) approach, Embedding Based approaches, and Partitioning Methods. XMTC utilizes the label hierarchy to partition extreme labels into multiple label groups, turning the extreme multi-label classification task into simple multi-group multi-label classification tasks. 
	
	Current research in multi-label classification encodes labels as a vector \cite{chalkidis2019large} The label vector ignores the semantic information of the labels. If the number of labels in each label group is different, then a classifier needs to be established for each label group. When the amount of label groups is large, an extremely large number of classifiers are required to build. How to build only one classifier, which can achieve multi-group multi-label classification tasks? If the extreme labels are enforced to be evenly grouped, the relations information among the labels in the hierarchy is missing. If you enforce the clip labels to balance the label group size, some labels’ information is missing.
	
	Therefore, this paper adopts the multi-question answering task to deal with the multi-label classification. This method can build only one classifier to train multi-label classification of label sets of different lengths. The approach does not sacrifice label information and inter-label relations. Furthermore, the method is theoretically applicable to zero-shot learning. Furthermore, this paper also proposed an auxiliary evaluation metric for classification in multi-answer questioning tasks. 
	
	According to the Natural Legal Language Processing (NLLP) community \footnote{\url{https://nllpw.org/}}, the automatic analysis of legal text is promising since in real life, legislation documents are in long size, with large volume, may be in different languages and are tagged with extreme multi-labels. And there are many other informal online texts with legal significance. Thus, NLLP suggests applying natural language processing on legal domain and to target on legal text analysis. This study adopts the proposed method and the evaluation metric to address the extreme multi-label classification problem in the legal domain. 
	
	Transformer-based pre-trained language models are utilized in text classification using feature engineering automatically alongside classification. \cite{minaee2021deep} states there are five steps for using pre-trained models: to select the suitable pre-trained model, to do domain adaptation, to design the task-specific model, to do task-specific fine-tuning, and to do model compression. The research also studies the task distribution in fine-tuning Bert-based pre-trained models in the legal domain. (The models and the results are shared at \footnote{\url{https://github.com/stuwangli/QA-Multilabel-EURLEX}})
	
 	\subsection{Contributions}
	
	Contributions of this paper are:	
	\begin{itemize}
		\itemsep0em
		\item This paper adopts the multi-answer questioning task for multi-label classification. This method can build only one classifier to train multi-label classification of label sets of different lengths. The approach does not sacrifice label information and inter-label relations. The method is theoretically applicable to zero-shot learning.
		\item Furthermore, this paper also proposed an auxiliary evaluation metric for classification in multi-answer questioning tasks.
		\item Proposed methods and evaluation metrics are adopted in the legal domain for extreme multi-label classification. Bert based legal domain pre-trained models are studied for multi-label classification tasks and the task distribution in utilizing pre-trained models is studied.
	\end{itemize}
	
	\section{Related Work}
	
	Transformer based pretrained models, such as BERT, achieved SOTA results in many NLP tasks. This paper selects BERT as the suitable pretrained model for multi-label classification. Huggingface provides the platform for developers to share their models, which contain domain adapted models. And there is several BERT based pretrained models in legal domain. Famous ones are legal BERT by \cite{chalkidis2020legal} and \cite{zheng2021does}. In legal domain adaptation in pretraining, \cite{chalkidis2020legal} try to prove the effectiveness of domain pretraining and \cite{zheng2021does} tries to answer when domain pre-training helps. The experiments show domain pretraining from both scratch and continual pre-training works competitive than BERT. however, domain pretraining on easy tasks performs similar as BERT. And domain pre-training helps difficult tasks more than easy tasks. However, current legal BERT models experiment the effectiveness of domain pretraining on the perspective of domain transfer, but not considering downstream tasks distribution which is relevant to task transfer. This paper adopted legal domain BERT from \cite{chalkidis2020legal} and \cite{zheng2021does} in the multi-label classification task to study the task distribution, also considering the negative transfer.  

	Directly using traditional MultiLabelBinarizer \footnote{\url{sklearn.preprocessing.MultiLabelBinarizer — scikit-learn 1.0.2 documentation }} and flat classification (\cite{silla2011survey} will encode extreme large labels into an extreme long vector. Computer requires large RAM to process a long vector. In addition, the label has meanings. MultiLabelBinarizer encoding erases the own semantic of these labels. Besides, different input encoder and label encoder will project inputs and labels into different latent spaces. To get the correlation between input and categories, the model better embeds them into the same latent space \footnote{\url{https://joeddav.github.io/blog/2020/05/29/ZSL.html}}.
	
	Referring to the MRPC dataset in GLUE \footnote{\url{https://huggingface.co/datasets/glue/viewer/mrpc/test}}, which compares the equivalent of sentence1 and sentence2, the multi-label classification tasks can change into the equivalent/similarity task between input and the label. And the label and the input are embedded into the same latent space. Apparently, changing multi-label classification into the similarity-based binary classification is good. But the dataset will be enlarged extremely as each input will be paired with each label. In extreme multi-label text classification, the training dataset could utilize thousands of labels. After transforming multi-label classification into binary classification, the amount of training dataset will be the original dataset size multiplying the label size. It will cost too long to train the dataset.
	The extractive question answering task \footnote{\url{https://huggingface.co/docs/transformers/tasks/question_answering}} targets on detecting the start index and end index of the answer span inside the text according to the question. One advantage of predicting index with start index and end index is the final answer is the span from start index to end index inside the text. Therefore, the length of the span could be dynamic according to index prediction. This is suitable for multi-word knowledge extraction tasks. The multi-label classification task has multi-words labels with dynamic length which takes advantage of the span extraction.
	
	However, the extractive question answering task only can predict one answer span from the text. But for XMTC, each text might have multiple labels. How to make an extractive model to do multiple spans extraction/multi-label classification? A method to handle multi-answer processing in the SQuAD dataset which filters the record with various answers in the training dataset and compares the predicted answer with all answers to select the one with the highest score in evaluation. \footnote{\url{ https://huggingface.co/course/chapter7/7?fw=pt}} The multi-answer records are small in the SQuAD dataset. But in most XMTC, documents have more than two concept labels. Therefore, the experiment cannot directly use filtering but requires a more proper way to do multi-spans extraction. 
	
	The multi-spans extraction problem is like the entity extraction problem. Therefore, according to the tutorial in Huggingface for entity extraction task \footnote{\url{https://huggingface.co/course/chapter7/2?fw=pt}}, the experiment also considers token level classification. The token level classification task is like a sequence tagging task. When applying the sequence tagging in multi-answer questioning, the output is a list of tags that present whether a token is the right category token or not. If there are more than one concept inside the categories list, more tokens are labeled as categories. In the question answering task, the question does not require labels. \cite{segal2019simple} also considers the multi-answer question problem as a sequence tagging problem, and only the answer is tagged with labels. Huggingface \footnote{\url{https://huggingface.co/course/chapter7/2?fw=pt}} mentions, if a token with label -100, it will not be considered in entropy loss calculation. Therefore, the experiment uses -100 to label sentence 1 and uses IO tagging, which performs better than BIO in multi-answer questioning \cite{segal2019simple} to label sentence 2.
	
	Let us open the black box and have a look at layers inside the BERT-base model. The BERT-base model contains BERT EMBEDDING and BERT ENCODER. The BERT ENCODER has 12 layers ranging from LAYER  0 ~ LAYER 11. After BERT ENCODER, there is a dropout layer, and the classifier layer follows the dropout. BERT has millions of parameters in these layers. \cite{lee2019would} stated that few layers, in specific, few bottom layers, which is closer to downstream task, in transformer-based models are affected most in fine-tuning. And the top layers extract linguistic syntax are more utilized to represent inputs. \cite{lee2019would}Lee et al. (2019) studied the specific number of layers affected in fine-tuning. From their experiment on BERT, they concluded that fine-tuning the last 3-5 layers in BERT base could achieve comparative performance of whole model fine-tuning. Therefore, in this experiment, the concept classification model freezes the first 11 layers and the EMBEDDING layers in BERT, with other layers inside the concept classification model unfrozen, and carries on fine-tuning on the built dataset. Other compression methods such as distillation will be discovered in future work.

	\section{Methodology}
	\begin{figure}[htp]
		\centering
		\includegraphics[width=3in]{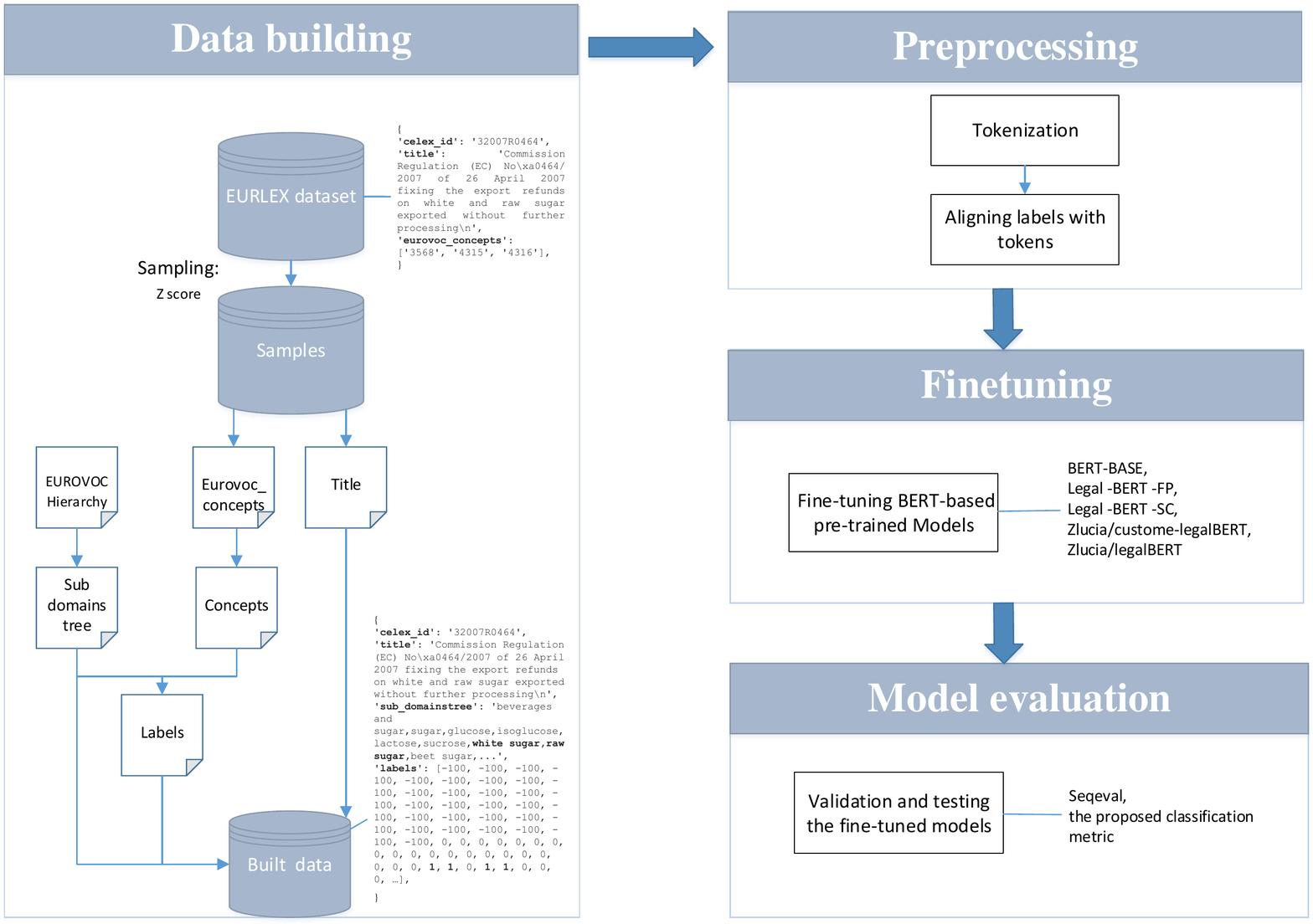}
		\vspace{-.4cm}
		\caption{Methodology framework}
		
	\end{figure}

	Figure 1 display the overall framework of the research methodology. The first step is data building to prepare dataset for the multi-answer questioning task. EURLEX dataset is sampled with Z-score to get the simulation data. The labeled eurovoc concepts are mapped to subdomain trees(/categories list) in EUROVOC hierarchy to get the labels for the multi-answers. Then labels are combined with title(/text) as the inputs for extractive multi-answer questioning task. Second step processes the inputs with tokenization and aligning labels with tokens. The pre-processed data are utilized in third step to fine-tune BERT-based pre-trained models for the multi-answer question task. And the performance of the fine-tuned models are evaluated with seqeval and the proposed auxiliary classification metric on validation and test samples. Subsections give the important components in the methodology. And the details of data building are stated in the next section.
	   
	\subsection{Label Partition with Label Hierarchy}
	According to \cite{silla2011survey}, the hierarchy classification includes building the hierarchy (e.g., creating new meta-classes) and classification according to the built hierarchy. With built hierarchy, the classification problem turns into a structure exploration problem. 
	
	The experiment uses the EURLEX dataset for extreme multi-label classification. These extreme large labels are concepts in EUROVOC. The EUROVOC \footnote{\url{https://eur-lex.europa.eu/browse/eurovoc.html}} is in a hierarchy tree structure. Except the geography domain, nodes on the tree are in inherit relations. The target classification labels, the concepts, are leaves in the tree. Therefore, the EUROVOC tree is almost the “IS-A” hierarchy and is suitable to be utilized directly for hierarchy classification. Although the geography domain, which contains nodes with same names, is the exception, it only affects the hierarchy classification on the geography domain. This issue will be considered in further requirements.  
	
	\subsection{Multi-label Text Classification Model}
	\begin{figure}[htp]
		\centering
		\includegraphics[width=3in]{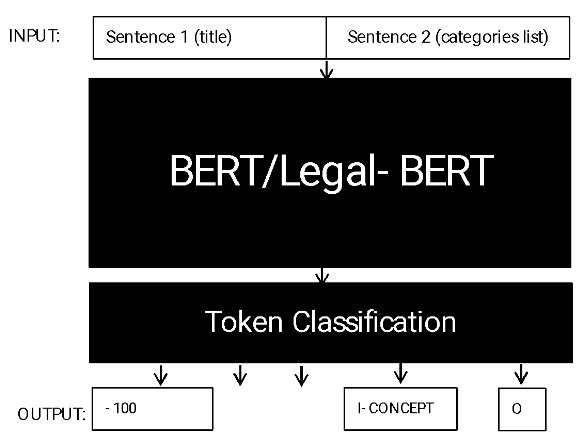}
		\vspace{-.4cm}
		\caption{The multi-label text classification model}

	\end{figure}
	
	The multi-label text classification can turn into an extractive multi-answer questioning problem and can be handled with a token classification task. Thus, shown in Figure 2, the multi-label classification model directly stacks a token classifier following the BERT-based pre-trained model. The input of the model is title(/text) (sentence 1) pads with categories list(/subdomain tree) (sentence 2) as one input sequence. Then the input sequence comes into a pretrained Bert-based model to generate its hidden representation. The hidden representation will be utilized as the input of the downstream token classifier to select labels from [‘I-CONCEPT’, ‘O’] for each token in the input.
	
	\begin{table*}[htp]
		\centering
	\begin{adjustbox}{max width=\textwidth}
		\begin{tabular}{llllll}
			\toprule
			Models   & Dataset & Pre-training Approaches & Initialized Model & Self-supervised Tasks & Released Links.  \\
			\midrule
			\textsc{BERT-base-uncased} \cite{devlin2018bert} & BookCorpus, English Wiki,  & Scratch & -  & MLM, NSP  & https://huggingface.co/bert-base-uncased 
			\\
			\textsc{LEGAL-BERT-FP} \cite{chalkidis2020legal} & EURLEX, ECHR, EDGAR, 12 G  & Continual pre-training & bert-base-uncased  & MLM  & https://archive.org/details/legal\_bert\_fp 
			\\
			\textsc{LEGAL-BERT-SC} \cite{chalkidis2020legal} & EURLEX, ECHR, EDGAR, 12 G  & Scratch & -  & MLM  & https://huggingface.co/nlpaueb/legal-bert-base-uncased 
			\\

			\textsc{zlucia/custom-legalbert}\cite{zheng2021does} & the CaseHOLD Dataset  & Scratch + domain-specific legal vocabulary & -  & MLM  & https://huggingface.co/zlucia/custom-legalbert  
			\\
			\textsc{zlucia/legalbert}\cite{zheng2021does} & the CaseHOLD Dataset  & Continual pre-training & bert-base-uncased  & MLM  & https://huggingface.co/zlucia/legalbert 
			\\
			\bottomrule
			
		\end{tabular}
	\end{adjustbox}
	\caption{BERT-based pre-trained models in legal domain}

	\end{table*}
	
	Table 1 displays pretrained models utilized for multi-label text classification. As known, transformer based pretrained models, such as BERT, achieved SOTA results in NLP tasks by simply fine-tuning step. And pre-trained models are suggested to do domain adaptation before fine-tuning. Table 1 shows the details of pretrained models in this methodology. As you can see, BERT \cite{devlin2018bert} is pre-trained with general knowledge such as Book Corpus and Wikipedia. Legal-BERT-SC and Legal-BERT-FP are legal domain specific models from \cite{chalkidis2020legal}. Legal-BERT-SC is pretrained from Scratch with 12 GB of English legal text, in specific, EURLEX, ECHR, EDGAR. Legal-BERT-FP is initialized by BERT and continually pre-trained with the same legal documents. zlucia/custom-legalbert and zlucia/legalbert are legal domain BERT from \cite{zheng2021does}. The zlucia/custom-legalbert is pre-trained from scratch with the CaseHOLD dataset combining with domain-specific legal vocabulary and the zlucia/legalbert is continual pre-trained with the same dataset without additional legal vocabulary. The legal BERT models with continual pre-training on domain knowledge are initialized on bert-base-uncased model. These models have been released in huggingface.co and   Archive.org. This research selects the Legal-BERT-FP/ bert-base-500k model for Legal-BERT-FP. The legal domain BERTs are trained from scratch and continual pre-trained in mask language modeling (MLM) self-supervised learning task. In this paper, BERT and BERT-BASE are both short names of BERT-base-uncased. zlucia/custom-legalbert is also mentioned as ZLUCIA/CUSTOM-LEGALBERT and Zlucia/custome-legalBERT. Similarly, zlucia/legalbert is called as ZLUCIA/LEGALBERT and Zlucia/legalBERT.

	\begin{table*}[htp]
		\centering
		\begin{adjustbox}{max width=\textwidth}
			\begin{tabular}{lll}
				\toprule
				Models   & Strengths & Weaknesses  \\
				\midrule
				\textsc{LEGAL-BERT-FP} & Adopting domain through continual pretraining.  & Not consider task distribution. 
				\\
				&Grid search impacts the performance.  & 
				\\
				&Performs better than BERT, especially in low resource test beds. & 
				\\
				&Adapt faster and better in ECHR cases, US contracts. &
				
				\\\midrule
				
				\textsc{LEGAL-BERT-SC} & Adopting domain through pre-training from scratch.  & Not consider task distribution. 
				\\
				&Grid search impacts the performance.  & Not consider the generalization ability in other domain tasks compared with continual pre-training models.
				\\
				&Performs better than BERT, especially in low resource test beds.  & 
				\\
				&Low perplexities (PPT) in EURLEX57K, ECHR-CASES, and CONTRACTS-NER. & 
				\\\midrule
				\textsc{zlucia/custom-legalbert}& Adopt domain through continual pretraining.  & Not consider task distribution, just considering prompt size in QA tasks. 
				\\
				&For the intermediate difficulty task that is not highly domain specific, domain pre-training can help & 
				\\
				&gain is most substantial for highly difficult and domain-specific tasks. &
				\\
				&Grid search impacts the performance. & 
				
				\\\midrule
				
				\textsc{zlucia/legalbert}& Adopt a domain with specific legal vocabulary and pre-training from scratch.   & Not consider task distribution just considering prompt size in QA tasks.
				\\
				&Domain pre-training helps in intermediate and difficult tasks.  & Not consider the generalization ability in other domain tasks comparing with continual pre-training model.
				\\
				&Gains better test performance than Legal BERT on Overruling, Terms of Service and CaseHOLD.&
				Grid search impacts the performance. 
				\\
				\bottomrule
				
			\end{tabular}
		\end{adjustbox}
		\caption{Comparison table for BERT-based pre-trained models in legal domain}
	\end{table*}
	
	Table 2 compares the strengths and weaknesses among the legal domain BERTs from \cite{chalkidis2020legal} and \cite{zheng2021does}. The domain pre-training from scratch and continual pre-training works more competitive than BERT. However, domain pretraining on easy tasks performs similar as BERT. And domain pre-training helps complex tasks more than easy tasks. However, current legal BERT models experiment the effectiveness of domain pretraining on the perspective of domain transfer, but not considering downstream tasks fine-tuning which is relevant to task transfer. This paper will study the effectiveness of legal domain BERTs in fine-tuning the multi-label classification task with different downstream task structures. Besides the proposed multi-answer questioning downstream task, the paper also studies the sequence classification and similarity tasks. Since the concept multi-label classification only be designed with multi-answer questioning task structure, the paper adds the domain multi-label classification which has constant domain categories list to study the relations between pre-trained models and different downstream tasks in fine-tuning. 
	
	\subsection{Evaluation Metrics}
	Huggingface \footnote{\url{https://huggingface.co/course/chapter7/2}} mentions that the evaluation of token classification utilizes the seqeval \footnote{\url{https://github.com/chakki-works/seqeval}} which is a Python framework for sequence tagging. The evaluation default mode is suitable for IO tags. And it supports metrics – precision, recall, f-score and accuracy. Since the seqeval tries to deal with chunking tasks, it is strict to identify the boundaries of the chunk. And the precision, recall and f-score are calculated based on mapping the exactly original chunks. 
	
	However, the extractive multi-label classification task focuses more on identifying the correct category instead of boundary detection. If only part tokens of the category label are extracted, the model still does well on the task because it pays attention to the right category. Therefore, the evaluation of extractive multi-label text classification is more likely to evaluate whether to extract information (e.g. tokens) from the right category instead of extracting the exact category chunk. Thus, the research proposes the other evaluation method on knowledge extraction by adding the category boundaries to the predicted category tokens before applying the seqeval. The similarity task utilizes the glue/mrpc metric\footnote{\url{ https://huggingface.co/course/chapter3/4?fw=pt}} for evaluation.

	\section{Dataset}
	
	Huggingface dataset gives a preview of EURLEX. The dataset contains four columns with document id, title, document text and the concept labels. The title is included inside the document text as the first paragraph. Each document has multiple concept labels tagged according to the EUROVOC list with only concept label id instead of tagging with the concept phrases. Most of these tagged concepts are showing in the title or their synonymous showing in the title. The longest document in EURLEX is above 20,000 tokens although most of the documents have less than 5,000 tokens. And few documents show having around 512 tokens, which is a suitable length for BERT input. Therefore, the experiment utilizes the title as the text for multi-label text classification. The dataset card mentions that EURLEX contains 45,000 training documents, 6000 validation documents and 6000 testing documents.
	
	There are thousands of concept categories inside the EURLEX dataset. If all the categories make one category list for the second sentence in the input, BERT meets the long input problem. Therefore, this experiment takes advantage of the EUROVOC hierarchy to split concept categories into groups. 
	
	The EUROVOC tree is downloaded from the EUROVOC website manually due to the dynamic web pages. The tree is stored locally and constructed by anytree package \footnote{\url{https://anytree.readthedocs.io/en/latest/}}. In tree construction, a root node named “eurovoc” is added as the ancestor of the tree. All node names are stored in lowercase for convenience to mapping concept labels. Including the root node, the tree height is 8. The first level is “eurovoc” and the second level are 21 domains in EUROVOC displaying in figure 3.
	
	\begin{figure}
		\centering
		\includegraphics[width=3in]{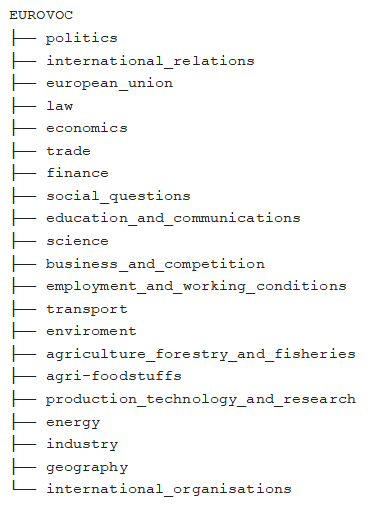}
		\vspace{-.4cm}
		\caption{Domains in EUROVOC}

	\end{figure}
	
	Since EUROVOC is a tree structure, nodes belong to different sub-trees with inner inherit relations (“IS-A” relations) or composition relations (“HAS-A” relations) between descendant nodes and parent nodes. Each sub-tree is like a domain cluster. A sub-tree starts from the parent node and ends with the leaf nodes linking the parent node directly and indirectly. If there are any nodes on the route from leaf and parent, they are also nodes belonging to the sub-tree. 
	Except the leaf node, every node on the hierarchy can be the parent node of a sub-tree. However, if the subtree is too large that the sub-tree contains too many nodes, the categories list will be too long to process. If the subtree is too small, the model requires a large amount of subtrees to cover all concept labels. Thus, the selection of subtrees should consider both scenarios.
	
	\begin{table*} [htp]
		\centering
		\begin{adjustbox}{max width=\textwidth}
			\begin{tabular}{ccccc}
				\toprule
				Parent Node Level   & Amount of Subtrees & Average Amount of Nodes in Subtrees & Maximum Amount of Nodes & Minimum Amount of Nodes  \\
				\midrule
				1 & 1  & 8274 & -  & -
				\\
				2 & 21  & 394 &  1597  & 137 
				\\
				3 & 127  & 65 & 532   & 1
				\\
				4 & 547  & 15 & 80   & 1  
				\\
				\bottomrule
				
			\end{tabular}
		\end{adjustbox}
		\caption{Analysis of EUROVOC hierarchy}

	\end{table*}
	
	Therefore, table 3 displays the analysis of the amount of subtrees and the amount of nodes in subtrees when selecting different levels as the parent node in EUROVOC hierarchy. Considering both scenarios, nodes on level 3 are utilized as the parent nodes of sub trees. The EUROVOC mentions that the concepts in EUROVOC are structured in 21 domains and 127 sub-domains. The level 3 parent nodes are these 127 sub-domains. Therefore, the research named the subtree of these subdomains as subdomain trees. 
	
% 	\begin{figure*}
% 		\centering
% 		\includegraphics[width=\textwidth]{researchmethodology.png}
% 		\vspace{-.4cm}
% 		\caption{multi-label text classification model}
% 	\end{figure*}
	To fine-tune the concept classification model, the EURLEX dataset utilizes subdomain trees to build the categories lists and labels concepts with [-100, I-CONCEPT, O] in sequence tagging. Figure 1 shows two columns in the EURLEX dataset are selected, in specific, the title and the concepts. Then the concepts are traversed on each subdomain tree and return the sequence labels for each pair of the title and the categories list. Considering the computation time cost to process the large built dataset, the experiment treats the task as a question-answer task without empty-answer samples by filtering the empty-answer samples. In other words, if the given concepts for the title are not in some subdomain trees, the pairs of the title and the categories lists are deleted from the built dataset. Here gives an example explaining the process.
	
	\begin{figure}
		\centering
		\includegraphics[width=3in]{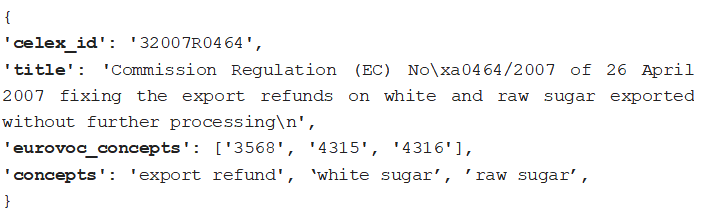}
		\vspace{-.4cm}
		\caption{Data record with ‘celex\_id’ of ‘32007R0464’}

	\end{figure}
	
	Figure 4 displays the record in EURLEX dataset. The "celex\_id" of the record is ‘32007R0464’. And the ‘eurovoc\_concepts’ are their concept ‘id’ inside the “eurovoc\_concepts.jsonl” file. To get the names/titles of the eurovoc\_concepts, the experiment firstly maps each ‘id’ to return its ‘title’ in the jsonl file. Therefore, the data record contains 3 concept ids, so the mapping returns 3 different concepts. 
	
	\begin{figure}
		\centering
		\includegraphics[width=3in]{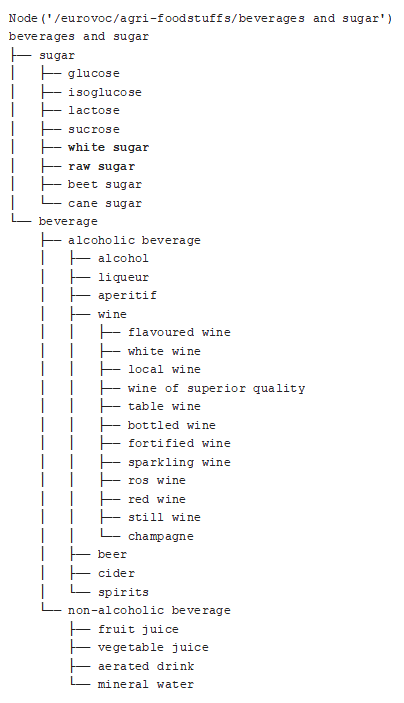}
		\vspace{-.4cm}
		\caption{Subdomain tree with parent node of ‘beverages and sugar’}

	\end{figure}
	  
	Then, each concept will be traversed inside the subdomain trees in EUROVOC. For example, in figure 5, the ‘white sugar’ and ‘raw sugar’ are on the subdomain tree with the parent node of ‘beverages and sugar’. 
	
	\begin{figure}
		\centering
		\includegraphics[width=3in]{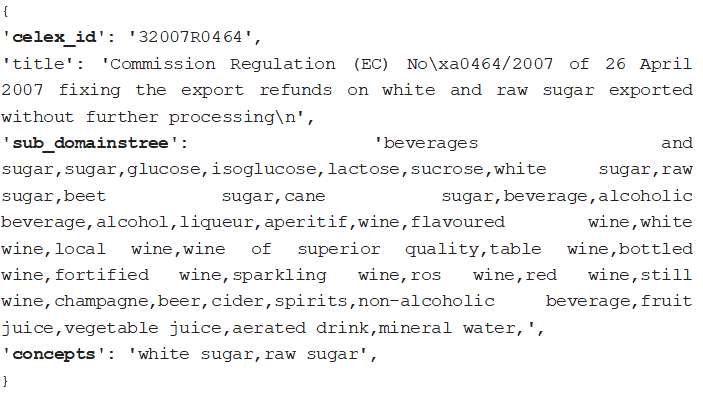}
		\vspace{-.4cm}
		\caption{The pair of ‘32007R0464’ title and ‘beverages and sugar’ subdomain tree}

	\end{figure}
	Figure 6 displays the categories list of the subdomain tree and the location of the ‘white sugar’ and ‘raw sugar’ inside the categories list. In the categories list, categories are isolated with commas which is easy for further sequence labeling. The data argumentation method by insertion of punctuation marks in the original text is also mentioned in \cite{karimi2021aeda}.
	
	\begin{figure}
		\centering
		\includegraphics[width=3in]{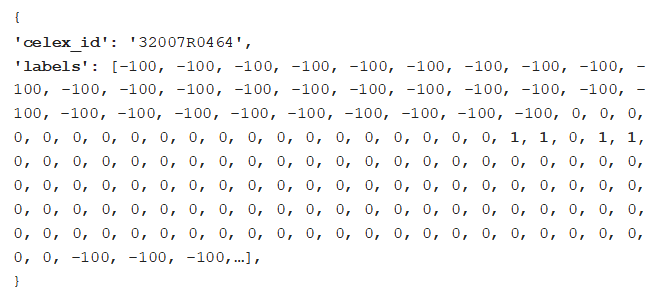}
		\vspace{-.4cm}
		\caption{The labels for the input with ‘32007R0464’ title and ‘beverages and sugar’ subdomain tree categories list}

	\end{figure}
	Figure 7 displays the sequence tagging labels for the input which is the pair of the ‘32007R0464’ title and the ‘beverages and sugar’ subdomain tree categories list. The title is tagged with -100 and the concepts in the categories list are tagged with 1 and others are tagged with 0. There are some more -100 after the categories list which are the labels of padded tokens.    
	
	\begin{table}[htp]
		\centering
		\begin{adjustbox}{max width=\textwidth}
			\begin{tabular}{llll}
				\toprule
				\   & Train & Validation & Test  \\
				\midrule
				Dataset  & 45000  & 6000 & 6000
				\\
				Built Dataset & 205053  & 27298 &  27149 
				\\
				Random Samples  & 381  & 362 & 362
				\\
				Built Samples & 1708  & 1650 & 1648  
				\\
				\bottomrule
				
			\end{tabular}
		\end{adjustbox}
		\caption{Dataset size}
	\end{table}

	Table 4 represents that after data building, the training dataset enlarges from 45000 into 205053. As the data is extremely large, the experiment randomly selects samples from the dataset and does experiments on the random samples. The sample selection is before the data building step. The sample sizes are calculated by an online sample size calculator \footnote{\url{https://www.qualtrics.com/experience-management/research/determine-sample-size/}} which is based on the Z-score. The confidence level is assigned with 95\%. Margin of error is assigned with 5\%. From the calculation, the train sample size is 381 from 45000. And the validation and test sample sizes are both 362 from 6000. After data building, the train, validation, and test data samples are 1708, 1650 and 1648. The random sample selection is referred to Hugging face \footnote{\url{https://huggingface.co/course/chapter5/3?fw=pt}} and Pandas \footnote{\url{https://pandas.pydata.org/docs/reference/api/pandas.DataFrame.sample.html}}.
	
	The multi-answer questioning task is first trained with negative samples. And one epoch lasts more than 20 hours for training. But the model performs really poorly in validation. Considering the computation time cost, the experiment treats the task as a question-answer task without empty-answer samples by filtering the empty-answer samples and training the dataset in 6 epochs. The understanding and utilizing of the negative samples is a problem and will be studied in future.
	
	\section{Evaluation Results and Discussion}
	
	\begin{table*}[t]
		\centering
		\hspace{-2mm}
		\resizebox{0.9\linewidth}{!}{
			\setlength{\tabcolsep}{1.3mm}
			\fontsize{9}{10}\selectfont
			
			\begin{tabular}{@{}lcrrrrrrrrrr}
				\toprule
				&&&& \multicolumn{4}{c}{Validation} &  \multicolumn{4}{c}{Testing}\\
				\cmidrule{5-8} \cmidrule{9-12} 
				\textbf{Models} & 
				\textbf{Epoch} & 
				\textbf{Training Loss} &
				\textbf{Validation Loss} &
				\multicolumn{1}{c}{\bf Precision} & 
				\multicolumn{1}{c}{\bf Recall} & 
				\multicolumn{1}{c}{\bf F-score} & 
				\multicolumn{1}{c}{\bf Accuracy} &
				\multicolumn{1}{c}{\bf Precision} & 
				\multicolumn{1}{c}{\bf Recall} & 
				\multicolumn{1}{c}{\bf F-score} &
				\multicolumn{1}{c}{\bf Accuracy}   \\
				\midrule
				\textsc{BERT-base-uncased} & 1 & 8.03&	4.65&	0.00&	0.00&	0.00&	99.01&	0.00&	0.00&	0.00&	98.99\\
				&2&4.38&	3.86&	15.65&	5.96&	8.63&	99.03&	15.01&	5.05&	7.55&	99.02\\
				&3&3.92&	3.58&	30.59&	15.85&	20.88&	99.10&	29.62&	14.07&	19.08&	99.08\\
				&4&3.67&	3.47&	32.53&	16.29&	21.71&	99.12&	31.66&	14.65&	20.03&	99.09\\
				&5&3.54&	3.40&	35.25&	20.55&	25.96&	99.13&	34.14&	18.58&	24.07&	99.11\\
				&6&3.44&	3.39&	35.38&	23.22&	28.04&	99.12&	34.59&	21.79&	26.73&	99.11
				\\\midrule
				
				\textsc{Legal-BERT-FP} & 1 & 8.81&	4.76&	0.00&	0.00&	0.00&	99.01&	0.00&	0.00&	0.00&	98.99\\
				&2&4.44&	3.87&	21.29&	3.83&	6.49&	99.03&	17.38&	2.96&	5.06&	99.02\\
				&3&3.79&	3.54&	31.73&	18.18&	23.11&	99.07&	30.45&	16.89&	21.72&	99.06\\
				&4&3.49&	3.33&	39.18&	17.55&	24.24&	99.14&	39.62&	17.13&	23.92&	99.12\\
				&5&3.33&	3.27&	41.93&	24.67&	31.07&	99.15&	40.42&	22.51&	28.92&	99.13\\
				&6&3.22&	3.27&	41.16&	27.29&	32.82&	99.15&	39.30&	25.13&	30.66&	99.13\\\midrule
				
				\textsc{Legal-BERT-SC} & 1 & 7.40&	5.06&	0.00&	0.00&	0.00&	99.01&	0.00&	0.00&	0.00&	98.99\\
				&2&5.01&	4.84&	0.00&	0.00&	0.00&	99.01&	0.00&	0.00&	0.00&	98.99\\
				&3&4.84&	4.66&	43.33&	0.63&	1.24&	99.02&	35.71&	0.73&	1.43&	99.01\\
				&4&4.68&	4.48&	35.51&	2.38&	4.45&	99.03&	40.60&	2.62&	4.92&	99.02\\
				&5&4.55&	4.39&	27.44&	3.54&	6.27&	99.02&	32.18&	4.08&	7.24&	99.02\\
				&6&4.46&	4.37&	26.87&	4.36&	7.51&	99.03&	30.15&	4.76&	8.21&	99.02\\\midrule
				
				\textsc{zlucia/custom-legalbert} & 1 & 6.72&	5.10&	0.00&	0.00&	0.00&	99.01&	0.00&	0.00&	0.00&	98.99\\
				&2&5.17&	4.97&	0.00&	0.00&	0.00&	99.01&	0.00&	0.00&	0.00&	98.99\\
				&3&5.07&	4.87&	0.00&	0.00&	0.00&	99.01&	0.00&	0.00&	0.00&	98.99\\
				&4&4.97&	4.80&	0.00&	0.00&	0.00&	99.01&	0.00&	0.00&	0.00&	98.99\\
				&5&4.94&	4.76&	0.00&	0.00&	0.00&	99.01&	0.00&	0.00&	0.00&	98.99\\
				&6&4.90&	4.74&	0.00&	0.00&	0.00&	99.01&	0.00&	0.00&	0.00&	98.99\\\midrule
				\textsc{zlucia/legalbert} & 1 & 7.83&	4.91&	0.00&	0.00&	0.00&	99.01&	0.00&	0.00&	0.00&	98.99\\
				&2&4.68&	4.18&	0.00&	0.00&	0.00&	99.00&	0.00&	0.00&	0.00&	98.99\\
				&3&4.15&	3.87&	14.08&	3.78&	5.96&	99.03&	12.21&	2.77&	4.51&	99.02\\
				&4&3.89&	3.69&	23.74&	7.56&	11.47&	99.07&	20.18&	5.58&	8.74&	99.05\\
				&5&3.71&	3.60&	28.57&	13.96&	18.76&	99.07&	25.85&	11.40&	15.82&	99.06\\
				&6&3.64&	3.57&	29.52&	14.15&	19.14&	99.08&	26.59&	11.55&	16.10&	99.07\\
				\bottomrule
			\end{tabular}
		}
		\vspace{-1mm}
		\caption{Validation and testing results (\%) in concept multi-label classification with multi-answer questioning downstream task with seqeval}
		%\label{tbl:sys_performance}
		% \vspace{-2mm}
	\end{table*}

	Table 5 shows validation and testing in concept multi-label classification. Since concept multi-label classification has various lengths of label groups, the downstream task utilized the multi-answer questioning task proposed in methodology, which has two sentences as an input and utilizes token classification for category tagging of sentence 2. The evaluation metrics include training loss, validation loss, and overall metrics - precision, recall, f1 score and accuracy in seqeval. 
	
	The training loss and validation loss in table 5 are all below 10\% from five pre-trained models. It shows that the BERT based models with freezing layers and multi-answer questioning downstream tasks are suitable to distribute the training and validation dataset. The concept multi-label classification task is so complex that the sentence 1 and sentence 2 in the input are both variables, and the task requires to predict labels for each token in the sentence 2. It requires enough parameters to simulate the task distribution. BERT and transformer-based models contain millions of parameters or more which are different from shallow models with small size. Large models are able to represent the complex task distribution.
	
	Validation and testing results display that the among five pre-trained models, Zlucia/custome-legalBERT cannot simulate the task with fine-tuning and other four models perform from good to bad are Legal-BERT-FP, BERT-base-uncased, and Zlucia/legalBERT. At the 6th epoch, Legal-BERT-FP gained about 5\% better than BERT-base-uncased in f-score, 6\% better in precision and 4\% better in recall. However, Zlucia/legalBERT is also a domain adapted pre-trained model with continual pre-training but achieves a lower f-score than BERT-base-uncased after fine-tuning. The difference between Legal-BERT-FP and Zlucia/legalBERT is that Zlucia/legalBERT is trained with CaseHOLD dataset with MLM task but Legal-BERT-FP is trained with EURLEX, ECHR, EDGAR dataset with MLM. 
	
	Legal-BERT-FP and Zlucia/legalBERT performs better than Zlucia/custome-legalBERT and Legal BERT-SC which shows that continual pre-training works better than training from scratch. Pre-training from scratch drops the pre-trained knowledge from BERT-base-uncased and taking advantage of BERT-base-uncased pre-trained knowledge can elaborate the performance. 
	
	Although Legal-BERT-FP performs better than BERT, their f-scores are close. It shows that BERT-base-uncased is a good few-shot learner that with fine/minor tuning, can gain similar performance as domain continual pre-trained BERT models.  
	
	The f-scores keep around 30\% and the precisions are below 50\% from both BERT and Legal-BERT-FP. To analyze the extracted concepts from Legal-BERT-FP at the 6th epoch, there are many empty extractions. It may be because the training sample size is too small to contain enough information to let the models learn. And most of the extracted concepts only contain part of tokens inside the category. Since the seqeval metric cares more on boundary detection, precision, the precision, recall and f-score are in low values. 
	
	The accuracy is high from the first epoch. However, at the first epoch, the precision, recall and f-score are all zeros. The reasons behind may be the samples are small and the labels have a minor percentage of label ‘I-CONCEPT’. Thus, if the models predicted all labels ‘O’ and due to the high percentage of label ‘O’ inside the labels, the accuracy was still high. The accuracy might show that there could be around 99\% ‘O’ labels in sequence tagging. 
	
	The results show the proposed multi-answer questioning model satisfies modeling the concept multi-label classification task. To gain better performance, the model requires more dataset in fine-tuning.
	
	Thinking as a human, when people do multiple choice questions, sometimes they can make the right selection from part of the answer instead of the whole answer chunk. Since the research focuses on the right category prediction instead of the span boundary detection, even though the model only identifies one token inside the right category, the model still works well in the task. Therefore, the evaluation should work to judge whether the model mines information or tokens from the right category. 
	
	To fulfill the evaluation expectation, the study proposes the second validation metric – classification metric, which adds the boundaries of categories according to the predicted tokens before applying the seqeval. As each category is separated by a ‘,’ in the category list, it is easy to locate the whole category label with a predicted token. It is similar as people do multi-choice questions, there are empties or symbols to separate different choices. 
	
	\begin{table*}[htp]
		\centering
		\hspace{-2mm}
		\resizebox{0.9\linewidth}{!}{
			\setlength{\tabcolsep}{1.3mm}
			\fontsize{9}{10}\selectfont
			
			\begin{tabular}{@{}lcrrrrrrrr}
				\toprule
				&& \multicolumn{4}{c}{Validation} &  \multicolumn{4}{c}{Testing}\\
				\cmidrule{3-6} \cmidrule{7-10} 
				\textbf{Models} & 
				\textbf{Epoch} & 
				\multicolumn{1}{c}{\bf Precision} & 
				\multicolumn{1}{c}{\bf Recall} & 
				\multicolumn{1}{c}{\bf F-score} & 
				\multicolumn{1}{c}{\bf Accuracy} &
				\multicolumn{1}{c}{\bf Precision} & 
				\multicolumn{1}{c}{\bf Recall} & 
				\multicolumn{1}{c}{\bf F-score} &
				\multicolumn{1}{c}{\bf Accuracy}   \\
				\midrule
				\textsc{BERT-base-uncased} & 1 & 0.00&	0.00&	0.00&	99.01&	0.00&	0.00&	0.00&	98.99\\
				&2&53.10&	19.53&	28.56&	99.01&	54.26&	17.61&	26.59&	99.01\\
				&3&61.81&	31.46&	41.70&	99.07&	63.72&	29.74&	40.56&	99.08\\
				&4&66.31&	32.62&	43.73&	99.12&	67.77&	30.81&	42.36&	99.11\\
				&5&62.86&	36.02&	45.79&	99.09&	65.21&	34.84&	45.41&	99.10\\
				&6&60.00&	38.68&	47.04&	99.06&	60.79&	37.46&	46.35&	99.07
				
				\\\midrule
				
				\textsc{Legal-BERT-FP} & 1 & 0.00&	0.00&	0.00&	99.01&	0.00&	0.00&	0.00&	98.99\\
				&2&58.15&	10.37&	17.61&	99.03&	60.76&	10.14&	17.38&	99.03\\
				&3&53.83&	30.34&	38.81&	99.02&	52.90&	28.77&	37.27&	99.00\\
				&4&68.54&	30.10&	41.83&	99.14&	68.03&	29.11&	40.77&	99.12\\
				&5&63.25&	36.45&	46.25&	99.11&	61.80&	33.67&	43.59&	99.08\\
				&6&61.60&	40.04&	48.53&	99.12&	59.69&	37.36&	45.96&	99.07
				
				\\\midrule
				
				\textsc{Legal-BERT-SC} & 1 & 0.00&	0.00&	0.00&	99.01&	0.00&	0.00&	0.00&	98.99\\
				&2&0.00&	0.00&	0.00&	99.01&	0.00&	0.00&	0.00&	98.99\\
				&3&96.55&	1.36&	2.68&	99.02&	90.00&	1.75&	3.43&	99.01\\
				&4&63.64&	4.07&	7.65&	99.01&	68.25&	4.17&	7.86&	99.01\\
				&5&51.38&	6.30&	11.23&	98.99&	58.94&	7.04&	12.57&	99.00\\
				&6&52.94&	8.29&	14.33&	99.00&	55.91&	8.49&	14.74&	98.99
				
				\\\midrule
				
				\textsc{zlucia/custom-legalbert} & 1 & 0.00&	0.00&	0.00&	99.01&	0.00&	0.00&	0.00&	98.99\\
				&2&0.00&	0.00&	0.00&	99.01&	0.00&	0.00&	0.00&	98.99\\
				&3&0.00&	0.00&	0.00&	99.01&	0.00&	0.00&	0.00&	98.99\\
				&4&0.00&	0.00&	0.00&	99.01&	0.00&	0.00&	0.00&	98.99\\
				&5&0.00&	0.00&	0.00&	99.01&	0.00&	0.00&	0.00&	98.99\\
				&6&0.00&	0.00&	0.00&	99.01&	0.00&	0.00&	0.00&	98.99
				
				\\\midrule
				
				\textsc{zlucia/legalbert} & 1 & 0.00&	0.00&	0.00&	99.01&	0.00&	0.00&	0.00&	98.99\\
				&2&0.00&	0.00&	0.00&	99.00&	0.00&	0.00&	0.00&	98.99\\
				&3&54.03&	13.96&	22.19&	99.03&	57.05&	12.57&	20.60&	99.03\\
				&4&61.35&	19.00&	29.02&	99.07&	64.56&	17.32&	27.31&	99.07\\
				&5&55.61&	26.66&	36.04&	99.02&	56.26&	23.97&	33.62&	99.02\\
				&6&58.38&	27.53&	37.42&	99.05&	60.48&	25.47&	35.85&	99.06
				
				\\
				\bottomrule
			\end{tabular}
		}
		\vspace{-1mm}
		\caption{validation and testing results (\%) in the concept multi-label classification for multi-answer questioning downstream task with proposed classification metric}
		%\label{tbl:sys_performance}
		% \vspace{-2mm}
	\end{table*}
	
	Table 6 shows the proposed classification metric evaluation results. The overall precision, recall and f-score are better than that in seqeval validation. The best precision is above 60\% and the f-score is near 50\%. The insertion of punctuation marks in the original text not only can help in data building but also can assist in classification task evaluation. 
	
	Validation and testing results indicate that Zlucia/custome-legalBERT cannot simulate the task with fine-tuning samples and other four models perform from good to bad are Legal-BERT-FP, BERT-BASE, and Zlucia/legalBERT. And Legal-BERT-FP and BERT-base-uncased gain competitive f-scores in validation and testing dataset. At epoch 5, Legal-BERT-FP performs slightly better than BERT in validation and slightly worse in testing. 
	
	The extracted answers are not independent of the positions of the category in categories list. The extracted answers might be different if the same title and same domain-tree randomly reorders the categories in the categories list. However, in classification, the order of the categories is not important. For example, the record in validation dataset with ‘celex\_id’ of ‘31999L0010’ extracts ‘foodstuff’ but gets ‘foodstuff’ and ‘processed foodstuff’ with a reordered categories list. \cite{chalkidis2019large} Ko et al. (2020) also mentions the position bias in extractive question answering. The other issue in the proposed multi-label classification task is the long input problem since BERT has limited input length.
	
	\begin{table*}[htp]
		\centering
		\hspace{-2mm}
		\resizebox{0.9\linewidth}{!}{
			\setlength{\tabcolsep}{1.3mm}
			\fontsize{9}{10}\selectfont
			
			\begin{tabular}{@{}lcrrrrrrrrrr}
				\toprule
				&&&& \multicolumn{4}{c}{Validation} &  \multicolumn{4}{c}{Testing}\\
				\cmidrule{5-8} \cmidrule{9-12} 
				\textbf{Models} & 
				\textbf{Epoch} & 
				\textbf{Training Loss} &
				\textbf{Validation Loss} &
				\multicolumn{1}{c}{\bf Precision} & 
				\multicolumn{1}{c}{\bf Recall} & 
				\multicolumn{1}{c}{\bf F-score} & 
				\multicolumn{1}{c}{\bf Accuracy} &
				\multicolumn{1}{c}{\bf Precision} & 
				\multicolumn{1}{c}{\bf Recall} & 
				\multicolumn{1}{c}{\bf F-score} &
				\multicolumn{1}{c}{\bf Accuracy}   \\
				\midrule
				\textsc{BERT-base-uncased} & 1 & 32.75&	26.10&	13.70&	7.99&	10.09&	88.75&	11.64&	6.86&	8.63&	88.26\\
				&2&24.80&	22.71&	38.09&	30.84&	34.08&	90.84&	34.67&	28.62&	31.35&	90.36\\
				&3&22.39&	21.40&	50.87&	44.77&	47.63&	91.49&	48.85&	43.18&	45.84&	91.19\\
				&4&21.27&	20.86&	52.60&	47.24&	49.78&	91.65&	51.57&	45.81&	48.52&	91.43\\
				&5&20.71&	20.46&	52.00&	46.39&	49.03&	91.87&	50.53&	44.71&	47.44&	91.63\\
				&6&20.21&	20.23&	52.87&	45.45&	48.88&	91.99&	51.32&	44.28&	47.55&	91.76
				
				\\\midrule
				
				\textsc{Legal-BERT-FP} & 1 & 35.09&	27.68&	26.52&	8.16&	12.48&	88.02&	25.07&	7.71&	11.79&	87.65\\
				&2&26.44&	25.18&	55.51&	33.81&	42.03&	88.34&	52.35&	32.09&	39.79&	87.86\\
				&3&24.78&	24.40&	43.01&	33.98&	37.97&	88.53&	39.67&	32.18&	35.53&	88.16\\
				&4&23.89&	23.47&	38.88&	35.34&	37.03&	89.44&	36.23&	33.53&	34.83&	89.10\\
				&5&23.10&	22.52&	44.64&	32.54&	37.64&	90.83&	40.63&	30.82&	35.05&	90.57\\
				&6&22.53&	22.16&	43.96&	33.98&	38.33&	90.97&	40.99&	32.35&	36.16&	90.84
				
				\\\midrule
				
				\textsc{Legal-BERT-SC} & 1 & 35.04&	27.77&	52.76&	16.23&	24.82&	88.12&	48.21&	14.82&	22.67&	87.68\\
				&2&26.38&	25.41&	55.06&	33.31&	41.50&	88.35&	52.14&	31.92&	39.60&	87.84\\
				&3&25.03&	25.02&	38.46&	33.98&	36.08&	88.14&	35.86&	31.67&	33.63&	87.61\\
				&4&24.49&	24.50&	37.85&	33.47&	35.53&	88.62&	34.74&	31.41&	32.99&	88.15\\
				&5&24.15&	23.90&	44.21&	31.44&	36.74&	89.80&	42.06&	29.81&	34.89&	89.31\\
				&6&23.77&	23.71&	43.50&	31.86&	36.78&	90.11&	40.02&	30.06&	34.33&	89.58
				
				\\\midrule
				
				\textsc{zlucia/custom-legalbert} & 1 & 36.80&	31.07&	0.00&	0.00&	0.00&	87.54&	0.00&	0.00&	0.00&	87.81\\
				&2&30.37&	27.33&	5.23&	2.04&	2.93&	88.10&	5.66&	2.29&	3.26&	87.75\\
				&3&28.03&	26.30&	39.41&	24.98&	30.58&	88.50&	35.14&	23.12&	27.89&	87.71\\
				&4&27.26&	26.06&	47.60&	32.80&	38.83&	88.49&	44.36&	30.65&	36.25&	87.76\\
				&5&27.13&	25.77&	48.10&	31.27&	37.90&	88.54&	44.75&	29.21&	35.35&	87.80\\
				&6&26.74&	25.79&	48.28&	33.39&	39.48&	88.47&	44.16&	31.08&	36.48&	87.73
				
				\\\midrule
				\textsc{zlucia/legalbert} & 1 & 33.89&	27.43&	0.28&	0.09&	0.13&	88.27&	0.28&	0.08&	0.13&	87.91\\
				&2&26.31&	24.71&	49.60&	15.89&	24.07&	88.57&	45.29&	14.65&	22.14&	88.21\\
				&3&24.13&	22.88&	44.00&	34.24&	38.51&	90.32&	41.50&	32.26&	36.30&	89.96\\
				&4&22.64&	21.78&	47.62&	40.02&	43.49&	90.98&	45.21&	39.12&	41.94&	90.68\\
				&5&21.77&	21.09&	47.44&	40.10&	43.46&	91.29&	45.34&	39.12&	42.00&	90.99\\
				&6&21.35&	20.82&	48.36&	40.19&	43.90&	91.47&	45.95&	38.95&	42.16&	91.21
				
				\\
				\bottomrule
			\end{tabular}
		}
		\vspace{-1mm}
		\caption{Validation and testing results (\%) in the domain multi-label classification for multi-answer questioning downstream task with seqeval}
		%\label{tbl:sys_performance}
		% \vspace{-2mm}
	\end{table*}

	Table 7 shows validation and testing in domain multi-label classification with the downstream task utilizing the multi-answer questioning task proposed in methodology. The domain multi-label classification has the constant domain label group which is different from concept multi-label classification having various lengths of label groups. The evaluation metrics include training loss, validation loss, and overall metrics - precision, recall, f1 score and accuracy in seqeval. Table 7 shows the validation results from the fine-tuned domain classification models, in specific, BERT-BASE, and four legal domain BERT. 
	
	The training loss and validation loss in table 7 are between 20\% to 40\% from five BERT based pre-trained models with freezing layers and multi-answer questioning downstream tasks. The domain multi-label classification task only has sentence 1 in the input as a variable and sentence 2 is a constant. Therefore, the complexity of the task is less than the concept multi-label classification task.
	
	Validation and testing results display that the five pre-trained models can simulate the domain multi-label classification task and perform from good to bad are BERT-base-uncased, Zlucia/legalBERT, Zlucia/custome-legalBERT, Legal-BERT-FP, and Legal BERT-SC. The continual pre-training is better than training from scratch. The BERT-BASE performs the best and the Zlucia/legalBERT and Zlucia/custome-legalBERT are better than Legal-BERT-FP, and Legal BERT-SC.   
	
	\begin{table*}[htp]
		\centering
		\hspace{-2mm}
		\resizebox{0.9\linewidth}{!}{
			\setlength{\tabcolsep}{1.3mm}
			\fontsize{9}{10}\selectfont
			
			\begin{tabular}{@{}lcrrrrrrrr}
				\toprule
				&& \multicolumn{4}{c}{Validation} &  \multicolumn{4}{c}{Testing}\\
				\cmidrule{3-6} \cmidrule{7-10} 
				\textbf{Models} & 
				\textbf{Epoch} & 
				\multicolumn{1}{c}{\bf Precision} & 
				\multicolumn{1}{c}{\bf Recall} & 
				\multicolumn{1}{c}{\bf F-score} & 
				\multicolumn{1}{c}{\bf Accuracy} &
				\multicolumn{1}{c}{\bf Precision} & 
				\multicolumn{1}{c}{\bf Recall} & 
				\multicolumn{1}{c}{\bf F-score} &
				\multicolumn{1}{c}{\bf Accuracy}   \\
				\midrule
				\textsc{BERT-base-uncased} & 1 & 53.08&	19.80&	28.84&	88.26&	49.66&	18.37&	26.82&	87.83\\
				&2&55.83&	39.85&	46.50&	88.55&	52.82&	38.10&	44.27&	88.02\\
				&3&60.10&	50.55&	54.91&	90.31&	58.75&	49.45&	53.70&	90.08\\
				&4&60.99&	52.34&	56.33&	90.74&	59.68&	50.89&	54.94&	90.63\\
				&5&61.92&	52.51&	56.83&	91.15&	60.18&	51.31&	55.39&	90.87\\
				&6&62.55&	51.66&	56.58&	91.37&	61.06&	50.72&	55.41&	91.13
								\\\midrule
				
				\textsc{Legal-BERT-FP} & 1 & 49.49&	16.40&	24.63&	87.99&	46.37&	15.16&	22.85&	87.58\\
				&2&54.14&	33.90&	41.69&	88.27&	50.94&	32.09&	39.38&	87.76\\
				&3&52.87&	40.70&	45.99&	88.14&	50.70&	39.71&	44.54&	87.73\\
				&4&52.64&	43.25&	47.48&	88.12&	50.83&	41.32&	45.59&	87.76\\
				&5&57.41&	38.83&	46.33&	89.58&	55.21&	38.10&	45.09&	89.19\\
				&6&56.70&	40.61&	47.33&	89.60&	55.41&	39.46&	46.09&	89.47
				\\\midrule
				
				\textsc{Legal-BERT-SC} & 1 & 50.26&	16.48&	24.82&	88.02&	47.37&	15.24&	23.06&	87.62\\
				&2&54.42&	33.47&	41.45&	88.28&	51.64&	32.01&	39.52&	87.80\\
				&3&49.62&	43.84&	46.55&	86.97&	47.85&	42.42&	44.97&	86.58\\
				&4&51.15&	43.33&	46.92&	87.45&	48.97&	42.08&	45.26&	86.99\\
				&5&55.05&	34.24&	42.22&	88.33&	52.25&	32.43&	40.02&	87.93\\
				&6&55.50&	36.02&	43.69&	88.52&	52.77&	34.72&	41.88&	88.12
				\\\midrule
				
				\textsc{zlucia/custom-legalbert} & 1 & 0.00&	0.00&	0.00&	87.52&	0.00&	0.00&	0.00&	87.79\\
				&2&50.65&	16.65&	25.06&	87.98&	45.72&	15.83&	23.52&	87.23\\
				&3&54.77&	32.20&	40.56&	88.27&	49.79&	30.57&	37.88&	87.56\\
				&4&52.49&	35.77&	42.55&	87.92&	49.82&	34.38&	40.68&	87.53\\
				&5&54.45&	34.32&	42.11&	88.24&	50.07&	32.18&	39.18&	87.59\\
				&6&52.68&	35.85&	42.67&	87.90&	49.33&	34.55&	40.64&	87.44
				\\\midrule
				
				\textsc{zlucia/legalbert} & 1 & 51.74&	16.40&	24.90&	88.06&	48.39&	15.24&	23.18&	87.65\\
				&2&51.42&	16.91&	25.45&	88.03&	48.08&	15.92&	23.92&	87.68\\
				&3&57.24&	41.63&	48.20&	89.68&	54.92&	40.64&	46.72&	89.36\\
				&4&58.40&	46.64&	51.87&	90.31&	56.59&	46.15&	50.84&	89.95\\
				&5&60.27&	48.60&	53.81&	90.85&	57.97&	47.76&	52.37&	90.49\\
				&6&60.99&	48.09&	53.78&	91.08&	58.70&	48.01&	52.82&	90.74
				\\
				\bottomrule
			\end{tabular}
		}
		\vspace{-1mm}
		\caption{Validation and testing results (\%) in the domain multi-label classification for multi-answer questioning task with proposed classification metric}
		%\label{tbl:sys_performance}
		% \vspace{-2mm}
	\end{table*}
	
	Table 8 shows the classification metric results in domain multi-label classification with the downstream task utilizing the multi-answer questioning task. Validation and testing results indicate that the five pre-trained models perform from good to bad are BERT-base-uncased, Zlucia/legalBERT, Legal-BERT-FP, Legal BERT-SC and Zlucia/custome-legalBERT. The BERT-base-uncased performs the best and the continual pre-training is better than training from scratch. And Zlucia/legalBERT is better than Legal-BERT-FP. Legal BERT-SC is better than Zlucia/custome-legalBERT.

	\begin{table*}[htp]
		\centering
		\hspace{-2mm}
		\resizebox{0.9\linewidth}{!}{
			\setlength{\tabcolsep}{1.3mm}
			\fontsize{9}{10}\selectfont
			
			\begin{tabular}{@{}lcrrrrrr}
				\toprule
				&&&& \multicolumn{2}{c}{Validation} &  \multicolumn{2}{c}{Testing}\\
				\cmidrule{5-6} \cmidrule{7-8} 
				\textbf{Models} & 
				\textbf{Epoch} & 
				\textbf{Training Loss} &
				\textbf{Validation Loss} &
				\multicolumn{1}{c}{\bf F-score} & 
				\multicolumn{1}{c}{\bf Accuracy} &
				\multicolumn{1}{c}{\bf F-score} &
				\multicolumn{1}{c}{\bf Accuracy}   \\
				\midrule
				\textsc{BERT-base-uncased} & 1 & 57.80&	48.56&	42.24&	7.73&	39.25&	9.94\\
				&2&44.98&	41.11&	43.52&	8.01&	40.79&	10.22\\
				&3&39.90&	38.05&	43.59&	8.01&	41.59&	10.22\\
				&4&37.54&	36.52&	44.12&	8.01&	41.14&	10.22\\
				&5&36.36&	35.78&	45.56&	8.29&	42.93&	10.22\\
				&6&35.76&	35.57&	45.29&	8.29&	42.41&	10.22
				
				\\\midrule
				
				\textsc{Legal-BERT-FP} & 1 & 61.34&	50.68&	44.38&	7.73&	42.69&	10.22\\
				&2&46.36&	41.50&	42.76&	8.01&	40.45&	10.22\\
				&3&40.04&	37.88&	42.90&	8.01&	40.53&	10.22\\
				&4&37.42&	36.34&	43.06&	8.01&	40.53&	10.22\\
				&5&36.26&	35.63&	43.14&	8.01&	40.60&	10.22\\
				&6&35.68&	35.41&	43.13&	8.01&	40.49&	10.22
				
				\\\midrule
				
				\textsc{Legal-BERT-SC} & 1 & 60.89&	50.65&	41.20&	5.25&	40.11&	4.70\\
				&2&46.61&	41.33&	42.64&	7.46&	38.79&	9.94\\
				&3&40.17&	37.95&	43.30&	8.01&	40.29&	10.22\\
				&4&37.65&	36.50&	43.13&	8.01&	40.33&	10.22\\
				&5&36.48&	35.80&	43.40&	8.01&	40.54&	10.22\\
				&6&35.97&	35.58&	43.33&	8.01&	40.51&	10.22
				
				\\\midrule
				
				\textsc{zlucia/custom-legalbert} & 1 & 53.09&	42.38&	43.40&	7.46&	41.69&	10.22\\
				&2&39.67&	36.94&	42.51&	8.01&	40.06&	10.22\\
				&3&36.26&	35.34&	42.26&	7.73&	40.17&	10.22\\
				&4&35.09&	34.75&	42.34&	7.73&	40.64&	10.22\\
				&5&34.75&	34.44&	42.69&	8.01&	40.83&	10.22\\
				&6&34.44&	34.36&	42.50&	7.73&	40.78&	10.22
				
				\\\midrule
				\textsc{zlucia/legalbert} & 1 & 61.10&	50.76&	43.00&	8.01&	40.26&	10.22\\
				&2&45.79&	41.05&	42.68&	8.01&	40.04&	10.22\\
				&3&39.54&	37.70&	43.42&	8.01&	39.91&	9.94\\
				&4&37.21&	36.34&	43.77&	8.01&	40.28&	10.22\\
				&5&36.17&	35.68&	43.57&	8.01&	40.32&	10.22\\
				&6&35.60&	35.46&	43.55&	8.01&	40.39&	10.22
				
				\\
				\bottomrule
			\end{tabular}
		}
		\vspace{-1mm}
		\caption{Validation and testing results (\%) in the domain multi-label classification for sequence classification downstream task}
		%\label{tbl:sys_performance}
		% \vspace{-2mm}
	\end{table*}
	Table 9 shows validation and testing in domain multi-label classification with the sequence classification downstream task. The sequence classification encodes the constant domain label group into a vector without utilizing the label semantic information. The evaluation metrics include training loss, validation loss, and f1 score, and accuracy. 
	
	The training loss and validation loss in table 9 are between 30\% to 60\% from five BERT based pre-trained models with freezing layers and sequence classification downstream task. The domain multi-label classification task only has sentence 1 as the input is a variable. Therefore, the complexity of the task is less than the multi-answer questioning task.
	
	Validation and testing results display that the accuracy is very low around 10\%. The five pre-trained models perform similarly in distributing the task and the BERT-base-uncased performs slightly better among them.
	
	\begin{table*}[htp]
		\centering
		\hspace{-2mm}
		\resizebox{0.9\linewidth}{!}{
			\setlength{\tabcolsep}{1.3mm}
			\fontsize{9}{10}\selectfont
			
			\begin{tabular}{@{}lcrrrrrr}
				\toprule
				&&&& \multicolumn{2}{c}{Validation} &  \multicolumn{2}{c}{Testing}\\
				\cmidrule{5-6} \cmidrule{7-8} 
				\textbf{Models} & 
				\textbf{Epoch} & 
				\textbf{Training Loss} &
				\textbf{Validation Loss} &
				\multicolumn{1}{c}{\bf Accuracy} & 
				\multicolumn{1}{c}{\bf F-score} &
				\multicolumn{1}{c}{\bf Accuracy} &
				\multicolumn{1}{c}{\bf F-score}   \\
				\midrule
				\textsc{BERT-base-uncased} & 1 & 30.71&	26.99&	90.92&	66.99&	90.95&	67.42\\
				&2&25.53&	29.19&	91.69&	68.30&	92.00&	69.66\\
				&3&21.44&	29.24&	92.00&	71.67&	92.61&	74.01
				
				\\\midrule
				
				\textsc{Legal-BERT-FP} & 1 & 33.97&	29.80&	88.65&	51.10&	88.36&	49.63\\
				&2&28.49&	29.00&	91.73&	66.95&	91.53&	66.42\\
				&3&25.21&	27.31&	92.07&	70.46&	92.17&	71.10
				
				\\\midrule
				
				\textsc{Legal-BERT-SC} & 1 & 43.17&	42.81&	84.52&	0.00&	84.46&	0.00\\
				&2&41.17&	43.58&	84.53&	0.17&	84.45&	0.00\\
				&3&42.25&	42.73&	84.53&	0.17&	84.45&	0.00
				
				\\\midrule
				
				\textsc{zlucia/custom-legalbert} & 1 & 40.81&	38.93&	85.04&	42.60&	84.48&	40.34\\
				&2&35.42&	35.68&	85.75&	46.47&	85.42&	45.26\\
				&3&34.24&	35.26&	85.19&	49.42&	85.08&	48.83
				
				\\\midrule
				\textsc{zlucia/legalbert} & 1 & 35.50&	34.90&	88.48&	44.42&	88.10&	43.40\\
				&2&31.24&	31.77&	89.36&	58.15&	89.40&	58.32\\
				&3&27.65&	31.49&	90.08&	60.11&	90.32&	61.51
				
				\\
				\bottomrule
			\end{tabular}
		}
		\vspace{-1mm}
		\caption{Validation and testing results (\%) in the domain multi-label classification for similarity downstream task}
		%\label{tbl:sys_performance}
		% \vspace{-2mm}
	\end{table*}
	
	Table 10 shows validation and testing in domain multi-label classification with the downstream similarity task. The evaluation metrics include training loss, validation loss, and overall metrics - f1 score and accuracy in glue/mrpc. 
	
	The training loss and validation loss in table 10 are between 20\% to 45\% from five BERT based pre-trained models with freezing layers and similar downstream tasks. The domain multi-label classification task has two variables, in specific, the sentence 1 and sentence 2 as the input. And the output is a binary classification for similarity or dissimilarity. 
	
	Validation and testing results display that the five pre-trained models can simulate the domain multi-label classification task and perform from good to bad are BERT-base-uncased, Legal-BERT-FP, Zlucia/legalBERT, Zlucia/custome-legalBERT, and Legal BERT-SC. The continual pre-training is better than training from scratch. The BERT-base-uncased is the best and the Legal BERT-SC is the worst with nearly 0\% f-score in validation and testing. The Legal-BERT-FP is better than Zlucia/legalBERT. 
	
	In conclusion, in fine-tuning a downstream task which is different from the task distribution of pre-training and continual pre-training tasks, BERT-BASE shows advantages in learning domain distribution and task distribution. However, the advantages of domain adapted BERTs are not apparent. And with different downstream task structures, different domain pre-trained BERT models perform differently. It is difficult to judge which domain BERT is better than others and it is difficult to get the reasons behind the scenarios. One reason might be that the negative task transfer makes learning difficult in fine-tuning.
	
	Comparing among four domain-adapted BERTs, even with domain adaptation, pre-trained models without continual on BERT-base-uncased pre-training knowledge has difficulty to learn downstream tasks in fine-tuning. 
	
	When the downstream task is more complex which takes full use of parameters in the pre-trained model, the training and validation losses are low and the task distribution is well learnt in fine-tuning. At that time, the advantages of domain continual pre-training display in Legal-BERT-FP which perform better than BERT, but not happening in Zlucia/legalBERT. And with the proposed classification metric, the advantages are not clear anymore. The effectiveness of continual pre-training should be verified in further works.    
	
	\section{Conclusion and Future Work}
	
	The classification problem is different from the extraction problem. The extraction task focuses on boundary detection. However, the classification problem is more like a selection problem. Inserting commas to separate categories in category list, the category boundary is easy to get. Therefore, if the prediction contains tokens in the right category, the model has made the right selection of the category. 
	
	\cite{karimi2021aeda} states that the insertion of punctuation marks into the input sequence could be an easier data augmented approach without losing the input information. And it helps in changing the positions of words in a sentence without disturbing the orders of the words which shows better generalization ability. However, future works should concern on selecting suitable symbols, for example, to use unique symbols from symbols inside the input text. If using context symbols, the coding will generate inaccurate data. 
	
	However, using incorrect built data, the fine-tuned models still gain similar conclusions. From checking, the amount of incorrect records is small. Whether a pre-trained model has the ability to ignore the incorrect input or task irrelevant inputs is still a question. 
	
	In previous review writing, there is an argument that continual pre-training helps narrowing transformer models into specific domain space which helps downstream domain specific task fine-tuning. Although the experiment shows Legal-BERT performs better than BERT in fine-tuning, it is still difficult to explain how it works since the evaluation results are close from both models. The advantage of continual pre-training is still not clear. Future works hope to explain the reasons inside.
	
	It is clear to see that fine-tuning with task specific data makes BERT and Legal-BERT achieve both domain knowledge and task knowledge. However, negative transfer happens without properly choosing pre-trained models for downstream tasks. Therefore, approaches to judge whether a pretrained model is suitable for the downstream task fine-tuning are required. The selection of downstream task implementations is close to the prompt tuning research. We  will look further on the topic to answer “how to select the prompt for the downstream task” combining with handling the problem “how to select a pre-trained model”. And future works will discover methods to make the judgment and how to judge simple and complex tasks. As shown in validation and testing results, Legal BERT performs not expected. Therefore, further works will look for the reasons behind the unexpected performances of domain-adapted pre-trained models.
	
	The reason for the wrong prediction might be that the title does not include enough concepts information for the domain labels, since the title is like the truncation of the legal document, which we discussed in a previous experiment report handling long input problems. 
	
	Previous experimental report mentions the label has meanings. It is good to enrich knowledge for labels, for example, to enrich domain knowledge for domain-driven labels. The EUROVOC shows the sibling nodes of domain nodes could be the knowledge to enrich the domain. Therefore, further experiments will take advantage of the EUROVOC hierarchy to select the layer information/node information as the enrichment knowledge, which aims to differentiate domains.  
	
	This experiment utilizes layer freezing to light the model. Other compression methods will be discovered in future work, such as the attention head distillation. 
	
	The reason behind the data size enlarging in building prompt data is how to make the negative samples. Extraction task makes both the positive and the negative answers in one input. Therefore, the extraction task is much easier in training and validation. However, extraction cannot perform as expected if without proper downstream tasks and if with long input. So the experiment requires a balance between the complexity of training and proper downstream task design. 
	
	The extracted answers depend highly on the positions of classes in the category list. The extracted answers are various from the same title and domain-tree which randomly reorders the labels in the domain tree for the multi-answer questioning task. In classification, the order of the labels is not important. \cite{chalkidis2019large} Ko et al. (2020) also mentions the position bias in extractive question answering. Future works need to consider how to deal with the position bias in the multi-answer questioning task for classification. 
	
	\cite{jurafskyspeech} state the information-retrieval-based and knowledge-based paradigms for question answering tasks. In IR-based methods, question type and other information assist the system to get retrievals. And the knowledge-based question answering focuses on mapping the query to structured data such as a database in order to retrieve relevant information inside the database. In this system, the answer is like an interface between the user interface and the system database. Like the user verification interface in the login system, the answer interface could be SQL code automatically mapping user queries to retrieve information from the database. Therefore, future works will focus on adding domain as a question type and look forward to semantic parsers for question answering. 
		
	\section*{Acknowledgement}
	We thank the all authors read and approved the final manuscript. 
	
	\bibliographystyle{IEEEtran}
	\bibliography{myref}

\end{document}